\documentclass[conference]{IEEEtran}
\IEEEoverridecommandlockouts

\usepackage{amsmath,amsfonts,bm}
\usepackage{mathtools}

\def\eqref#1{equation~\ref{#1}}

\def\1{\bm{1}}
\newcommand{\train}{\mathcal{D}}

\def\ve{{\bm{e}}}

\def\vw{{\bm{w}}}
\def\vx{{\bm{x}}}

\def\mW{{\bm{W}}}

\DeclareMathAlphabet{\mathsfit}{\encodingdefault}{\sfdefault}{m}{sl}
\SetMathAlphabet{\mathsfit}{bold}{\encodingdefault}{\sfdefault}{bx}{n}

\def\gS{{\mathcal{S}}}

\def\gV{{\mathcal{V}}}

\def\sD{{\mathbb{D}}}

\def\sR{{\mathbb{R}}}

\DeclareMathOperator*{\argmin}{arg\,min}

\newcommand{\nll}{\mathrm{NLL}}
\newcommand{\mse}{\mathrm{MSE}}
\newcommand{\qaft}{\textnormal{QAFT}}
\newcommand{\gptq}{\textnormal{GPTQ}}
\newcommand{\rtn}{\textnormal{RTN}}

\makeatletter
\renewcommand{\paragraph}{%
  \@startsection{paragraph}{4}%
  {\z@}{0.25ex \@plus 0.25ex \@minus .5ex}{-1em}%
  {\normalfont\normalsize\bfseries}%
}
\makeatother

\newcommand{\INT}[1]{\texttt{int{#1}}}
\newcommand{\model}[1]{\texttt{#1}}

\usepackage{cite}
\usepackage{amsmath,amssymb,amsfonts}
\usepackage{algorithmic}
\usepackage{graphicx}
\usepackage{textcomp}
\usepackage{xcolor}
\usepackage[numbers]{natbib}
\usepackage{afterpage}
\usepackage{cuted}
\usepackage{multicol}
\usepackage{mathtools}
\usepackage{hyperref}
\def\BibTeX{{\rm B\kern-.05em{\sc i\kern-.025em b}\kern-.08em
    T\kern-.1667em\lower.7ex\hbox{E}\kern-.125emX}}
\begin{document}

\makeatletter
\newcommand{\linebreakand}{%
  \end{@IEEEauthorhalign}
  \hfill\mbox{}\par
  \mbox{}\hfill\begin{@IEEEauthorhalign}
}
\makeatother

\title{Understanding the Difficulty of Low-Precision Post-Training Quantization for LLMs}

\author{
\IEEEauthorblockN{1\textsuperscript{st} Zifei Xu}
\IEEEauthorblockA{
\textit{d-Matrix}\\
Santa Clara, USA \\
\texttt{xuzifei@d-matrix.ai}}
\and
\IEEEauthorblockN{2\textsuperscript{nd} Sayeh Sharify}
\IEEEauthorblockA{
\textit{d-Matrix}\\
Santa Clara, USA \\
\texttt{sayehs@d-matrix.ai}}
\and
\IEEEauthorblockN{3\textsuperscript{rd} Wanzin Yazar}
\IEEEauthorblockA{
\textit{d-Matrix}\\
Santa Clara, USA \\
\texttt{wyazar@d-matrix.ai}}
\linebreakand
\IEEEauthorblockN{4\textsuperscript{th} Tristan Webb}
\IEEEauthorblockA{
\textit{d-Matrix}\\
Santa Clara, USA \\
\texttt{twebb@d-matrix.ai}}
\and
\IEEEauthorblockN{5\textsuperscript{th} Xin Wang}
\IEEEauthorblockA{
\textit{d-Matrix}\\
Santa Clara, USA \\
\texttt{xwang@d-matrix.ai}}
}

\maketitle

\begin{abstract}
Large language models of high parameter counts are computationally expensive, yet can be made much more efficient by compressing their weights to very low numerical precision.  
This can be achieved either through post-training quantization by minimizing local, layer-wise quantization errors, or through quantization-aware fine-tuning by minimizing the global loss function.  
In this study, we discovered that, under the same data constraint, the former approach nearly always fared worse than the latter, a phenomenon particularly prominent when the numerical precision is very low.  
We further showed that this difficulty of post-training quantization arose from stark misalignment between optimization of the local and global objective functions.  
Our findings suggested limited utility in minimization of local quantization error and the importance of direct quantization-aware fine-tuning, in the regime of large models at very low precision.
\end{abstract}

\section{Introduction}

Large language models (LLMs) have become remarkably powerful and are increasingly deployed in real-world applications in recent years~\citep{xu2023paradigm, wang2024weaver, zheng2023building, wu2023brief}.  
Despite their effectiveness, LLMs are typically trained with weights in \texttt{float16}, making post-training compression techniques such as quantization crucial for efficient inference~\citep{yao2022zeroquant, kim2023squeezellm, kim2023finequant, park2022lut, frantar2022gptq, lin2023awq}. 

Two distinct methods are prevalent for LLM weight quantization. The first one, \emph{quantization-aware fine-tuning} (QAFT) optimizes a differentiable global objective just like at pretraining time, with quantization operations acting on the model weights, requiring backpropagation and gradient updates~\citep{chee2024quip, dettmers2023spqr, dettmers2024qlora, hu2021lora, li2023loftq, huang2024billm}.  
The second method minimizes layer-wise local quantization errors, with no backpropagation needed~\citep{frantar2022optq, xiao2023smoothquant, lee2024owq, lin2023awq}.   
Both methods typically require a small amount of data and should theoretically achieve similar result since minimizing the local losses should in turn minimize the global loss, and \emph{vice versa}. 

Let us first introduce some formal notations.  
Denote the LLM network by $f_{\mW}: \sD \to \sR^{|\gV|}$, which takes text input $\vx \in \train \subset \sD$ and generates logits $f_{\mW}(\vx)$ across vocabulary $\gV$. 
Assume it is well-pretrained on language-modeling objective, \emph{i.e.} $\mW = \argmin_{\mW'} \nll(\vx|f_{\mW'})$ on a training data set $\train$.  
Here $\mW \triangleq \left(\mW_1, \cdots, \mW_L\right)$ collects all layer-wise weights $\mW_l$ with $l \in \{1, \cdots, L\}$ indexing layers.  
In the following we will also use vector notation $\vw \in \sR^D$ to represent an equivalent, flattened version of $\mW$ in the $D$-dimensional weight space.  
We denote $Q: \sR^D \to \sR^D$ as the weight (fake-)quantization function (for procedural details see Section~\ref{methods:quant}). 
We refer to the fake-quantized weights as \emph{round-to-nearest} ($\rtn$), $\mW_\rtn \triangleq Q(\mW)$.

$\qaft$ methods essentially keep optimizing the global objective with quantization in the loop, \emph{i.e.}
\begin{align}
    \mW_\qaft &= \argmin_{\mW'} \nll(\vx|f_{Q(\mW')}), 
\end{align}
where $\vx \in \train$; due to the non-differentiability of $Q(\cdot)$, straight-through estimator is commonly used in gradient back-propagation.  

In contrast, layer-wise quantization error minimization techniques seek to solve $L$ distinct layer-wise optimization problems:
\begin{align}
    \mW_l &=  \argmin_{\mW'_l} \mse\left(Q(\mW'_l) \vx_l, \mW_l \vx_l\right) \\
    &=  \argmin_{\mW'_l} \lVert Q(\mW'_l) \vx_l - \mW_l \vx_l \rVert^2 , 
\end{align}
where $\vx_l$ is the input to the $l$-th layer when $\vx \in \train$ is passed through the network $f$. 
One popular method of this kind, namely $\gptq$~\citep{frantar2022gptq}, derived from \emph{optimal brain compression} (OBC)~\citep{frantar2022optimal}, seeks to solve the layer-wise mean squared error ($\mse$) minimization problem by finding a less steeply rising direction in this local quadratic loss landscape, through efficient computation of the Hessian of the local $\mse$ losses.  
We denote a $\gptq$-optimized network's weights by $\mW_\gptq$.

While intuitively, both approaches should produce well-generalizing quantized networks, our systematic study reveals a misalignment. Local loss minimization often leads to suboptimal global loss, and \emph{vice versa}, especially at low quantization precisions. Through loss-landscape analysis, we offered an explanation of why the post-training quantization by local loss minimization struggles to produced well-generalizing quantized networks, guiding future LLM quantization practices.

\section{Related work}
\paragraph{Post-training quantization}

Quantization reduces the memory and computational demands of neural networks by converting weights or activations from full precision to lower precision formats, like 8-bit integers. Post-training quantization (PTQ) techniques achieve this without retraining the network. Several PTQ techniques have emerged to reduce deployment cost of LLMs. Some methods focus on identifying outlier features that are difficult to quantize and either represent them with a higher precision, \emph{e.g.} LLM.int8()~\citep{dettmers2208llm}, or mitigate their quantization error by adding additional operations to the network, such as SmoothQuant~\citep{xiao2023smoothquant}, AWQ~\citep{lin2023awq}, and OWQ~\citep{lee2024owq}.

Other PTQ methods employ adaptive rounding techniques to reduce quantization errors. For instance, OBC~\citep{frantar2022optimal} quantizes weights one-by-one in a specific order based on the approximate second-order information of the weights, and adjusts the remaining weights to minimize the quantization error. GPTQ~\citep{frantar2022gptq}, also known as OPTQ~\citep{frantar2022optq}, extends OBC by enabling parallel quantization of weight matrices, applying the same quantization order to all rows of the weight matrix. Similarly, QuIP~\citep{chee2024quip} uses adaptive rounding to minimize a quadratic proxy objective of the quantization error.

\paragraph{Quantization-aware fine-tuning}
Fine-tuning LLMs ensures task-specific adapdations but is computationally expensive. Parameter efficient fine-tuning (PEFT) reuses some of the pretrained model’s parameters and selectively fine-tune a subset of parameters for the downstream tasks. Common PEFT methods include LoRA~\citep{hu2021lora}, QLoRA~\citep{dettmers2024qlora}, L4Q~\citep{jeon2024l4q}, LoftQ~\citep{li2023loftq}, Prefix and Prompt Tuning~\citep{li2021prefix, lester2021power, qin2021learning, liu2023gpt}, IA3~\citep{liu2022few}, and PEQA~\citep{kim2024memory}.

LoRA, QLoRA, L4Q, and LoftQ freeze pretrained model parameters and fine-tune on inserted task-specific adapters. Adapters undergo low rank decomposition to further reduce the trainable parameters~\citep{hu2021lora, dettmers2024qlora, jeon2024l4q, li2023loftq}.
In prefix and prompt tuning, the parameters of an original large language model are frozen, and only the trainable prompt embeddings are fine-tuned~\citep{li2021prefix, lester2021power, qin2021learning, liu2023gpt}. Similarly, in IA3 only the hidden state parameters are fine-tuned~\citep{liu2022few}. 

PEQA is a memory-efficient fine-tuning method for quantized LLMs that updates only the quantization scale, keeping the integer matrix frozen~\citep{kim2024memory}.

\paragraph{Ultra-low precision pretraining} 
Recent efforts aim to binarize LLMs by quantizing them to ultra-low bit-widths. For instance, PB-LLM~\citep{shang2023pb} and SpQR~\citep{dettmers2023spqr} employ a mixed-precision quantization technique, representing the majority of the weights with a single bit while retaining a small portion of the weight in the original high precision or \INT{8}. BiLLM~\citep{huang2024billm} utilize Hessian information to identify salient and non-sailents weights, employing binary residual approximation of salient weights and grouped quantization of non-salient weights. BitNet~\citep{wang2023bitnet} replaces transformer linear layers by a binary linear layer, retaining other components in high-precision. BitNet b1.58~\citep{ma2024era} is an extension of BitNet that utilizes ternary quantization for its weights, achieving better accuracy in downstream tasks compared to BitNet.

\paragraph{Loss landscape analysis} 
Analysis of deep neural networks' loss landscape has long been a tool toward understanding of the generalization properties of the optimized model~\citep{li2018visualizing,ghorbani2019investigation,sagun2017eigenvalues,sagun2018empirical}, as well as in explaining difficulties arising in efficient network optimization processes~\citep{evci2020difficulty}. 

\section{Methods}
\label{sec:methods}

\subsection{Models and data set}

We experimented with $11$ models from $3$ model families, namely GPT-2~\citep{radford2019language}, OPT~\citep{zhang2022opt} and Llama 2~\citep{touvron2023llama}.
All models were served by the Hugging Face Model Hub~\footnote{All models were accessed from official repositories hosted by \texttt{\url{https://huggingface.co/}} in April 2024.}.   

We used the WikiText-2~\citep{merity2016pointer} data set in all experiments.
Unless noted otherwise, the training split used in all procedures was of $128$ examples, each of the maximum sequence length supported by the model being experimented.  
Entire split was used for validation and test data.  

\subsection{Numerical data types and quantizer calibration}
\label{methods:quant}

We experimented with $5$ integer data types for weight quantization, namely \INT{8}, \INT{6}, \INT{4}, \INT{3} and \INT{2}, of varied numerical precision. 
For integer with $B$-bit precision, encoding range is symmetric, \emph{i.e.} $\{-2^{B-1}+1, \cdots, 2^{B-1}-1\}$, excluding $-2^{B-1}$; for example, \INT{2} quantization is effectively ternary, $\{-1, 0, 1\}$.  

We used PyTorch's quantization API~\footnote{See~\texttt{\url{https://pytorch.org/docs/stable/quantization.html}}.} to obtain the weight fake-quantization functions $Q(\cdot)$.
The quantization scheme used was per-tensor symmetric~\footnote{Note that different quantization schemes and/or different data types of the same precision often result in different generalization quality of the quantized network, determined by the granularity of a channel/group/block-wise scheme.  Here we choose to use the simplest scheme in order to conduct a controlled scientific study with fewest confounding factors. }. 
The scaling factor $a_l$'s of the fake-quantizer was determined by mean squared quantization error minimization, \emph{i.e.} $a_l = \argmin_a \lVert Q_a(\mW_l) - \mW_l \rVert^2, \forall l \in \{1,\cdots\,L\}$, by means of PyTorch's \texttt{HistogramObserver} mechanism.
For simplicity, only weights of layers in the transformer stack were quantized, sparing other weights such as those in embedding and prediction head layers.  

\begin{figure*}[!thb]
\centering
\includegraphics[scale=0.8]{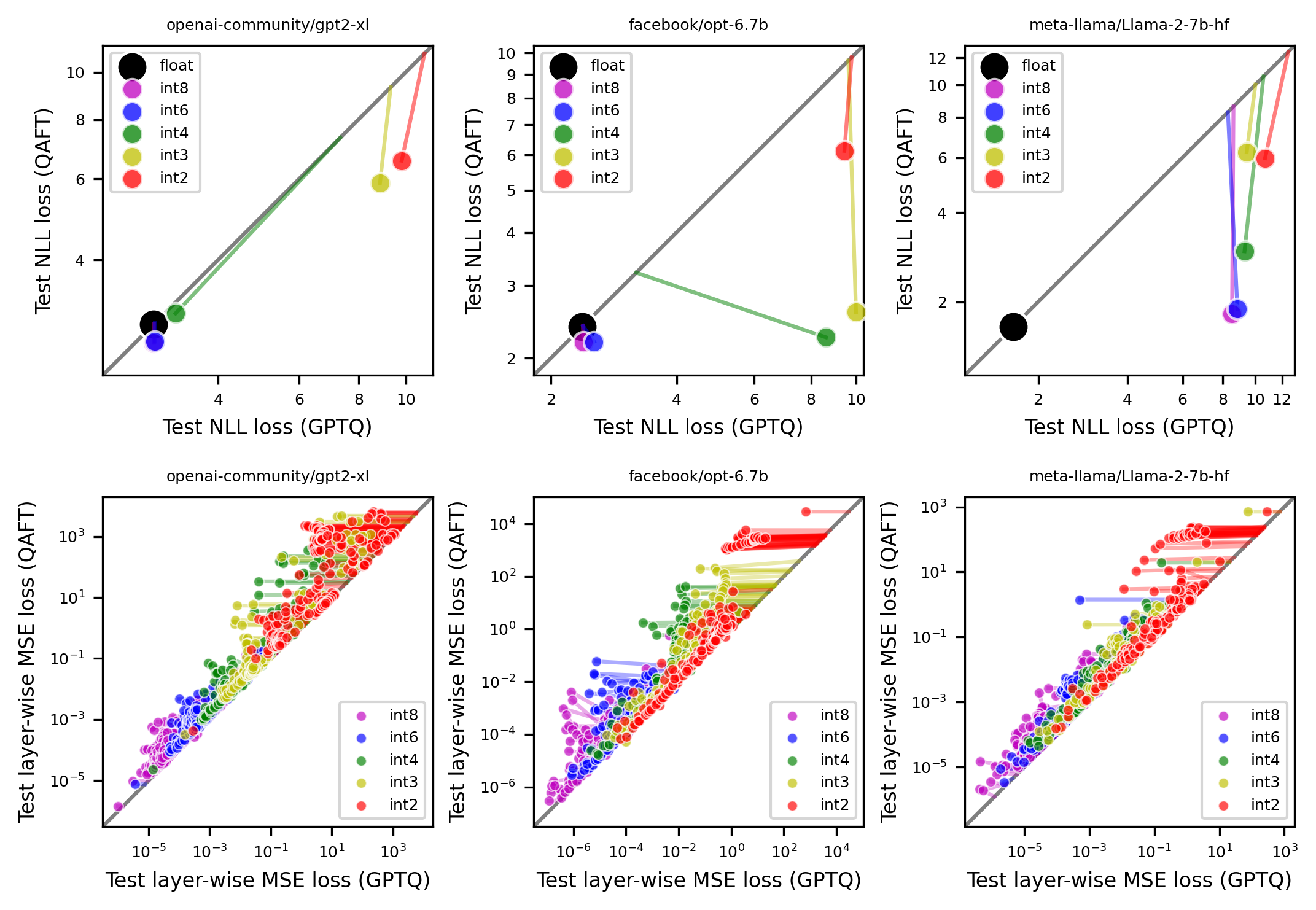}
\caption{
\textbf{Misalignment between minimization of the global $\nll$ loss ($\qaft$) and minimization of the local layer-wise $\mse$ losses ($\gptq$)}.  
The upper row shows global NLL losses, and the lower row presents layer-wise MSE losses for three models (one per column). Data points compare QAFT (vertical axis) to GPTQ (horizontal axis). The gray diagonal indicates identity. Black dots (if present) represent full-precision models, while colored dots mark losses after QAFT and GPTQ. Colored lines originating from each dot intersect the diagonal, showing RTN-quantized model losses for the corresponding format. In the lower row, symbols represent individual quantized layers.
}
\label{fig:gptq_vs_qaft}
\vspace{0.5cm}
\end{figure*}


\subsection{$\gptq$}

We followed the original GPTQ procedure~\citep{frantar2022gptq} exactly except for one change described below.  
As reported by other adopters of GPTQ, \emph{e.g.} \texttt{\url{https://huggingface.co/TheBloke}}, different choices of the dampening factor as a hyperparameter in the GPTQ procedure could lead to outcomes of varied qualities. 
In order to eliminate this confounding factor, we performed hyperparameter tuning for the GPTQ dampening factor over a search space of $\{10^{-3}, \cdots, 10^4\}$, in a layer-wise manner. 
As mentioned above, $128$ examples from the training split were used for the GPTQ procedure, consistent with the original work~\citep{frantar2022gptq}.  


\subsection{Quantization-aware fine-tuning ($\qaft$)}

To perform QAFT, the exact same $128$ training examples were used as above, for a fair comparison. 
We ran QAFT for $8$ epochs, \emph{i.e.} in total $1024$ training iterations (see Appendix~\ref{app:iterations} for a study on the effect of number of iterations). 
Straight through estimator (STE)~\citep{hinton2012deep, bengio2013estimating} was employed in back-propagation through quantization functions. 
Only quantized weights in the transformer network were subject to gradient updates. 
We used the AdamW optimizer~\citep{loshchilov2017decoupled} without weight decay, and with a linear learning rate decay schedule of $1$ order of magnitude.  
We ran a hyperparameter grid search over $4$ initial learning rate values,  $\{10^{-6},10^{-5},10^{-4},10^{-3}\}$ for all models except for \model{Llama-2-7b-hf}, in which case it was $\{10^{-6},10^{-3}\}$ (see Appendix \ref{app:lr} for more details on hyperparameter tuning); best $\nll$ loss over the validation split was chosen.  





\section{Results}

\begin{figure*}[!thb]

\centering
\includegraphics[scale=0.81]{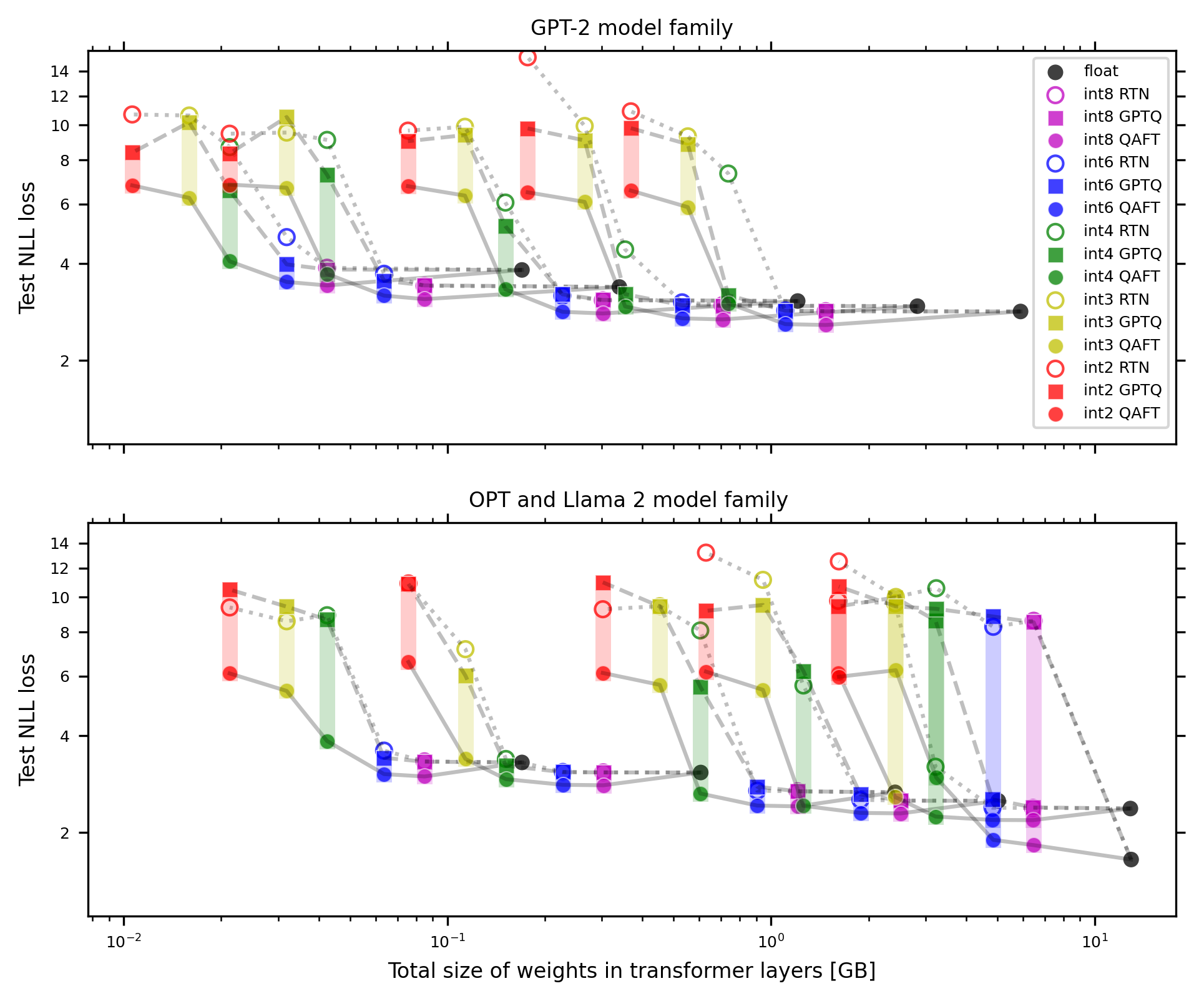}
\caption{
\textbf{Tradeoff between quantized model generalization and its weight size. }
\textbf{Upper}: models from the GPT-2 model family: \model{distilgpt2}, \model{gpt2}, \model{gpt2-medium}, \model{gpt2-large} and \model{gpt2-xl}. 
\textbf{Lower}: models from the OPT and Llama 2 families: \model{opt-250m}, \model{opt-350m}, \model{opt-1.3b}, \model{opt-2.7b}, \model{opt-6.7b} and \model{Llama-2-7b-hf}.
Black circles represent the full-precision models. 
Hollow colored circles are $\rtn$-quantized models, solid colored circles are $\qaft$-quantized models, and solid colored squares are $\gptq$-quantized models. 
Dotted, dashed and solid gray lines connect quantized solutions from the same model produced by $\rtn$, $\gptq$ and $\qaft$, respectively.
We highlight the difference between $\gptq$- and $\qaft$-quantized models with colored, transparent, vertical strips, for each quantized model.  }
\label{fig:tradeoff}
\vspace{0.5cm}
\end{figure*}

\begin{figure*}

\centering

\includegraphics{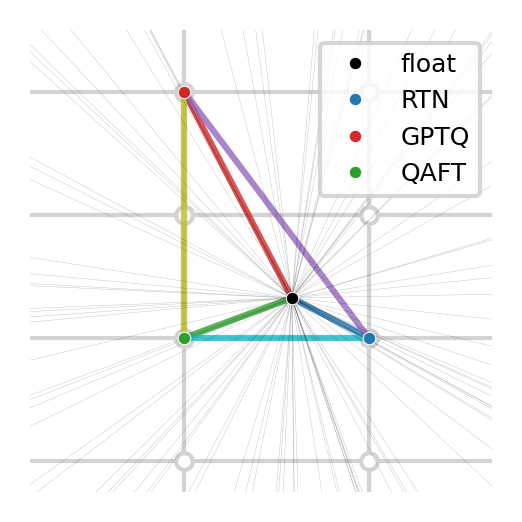} \\
\begin{multicols}{2}
\includegraphics{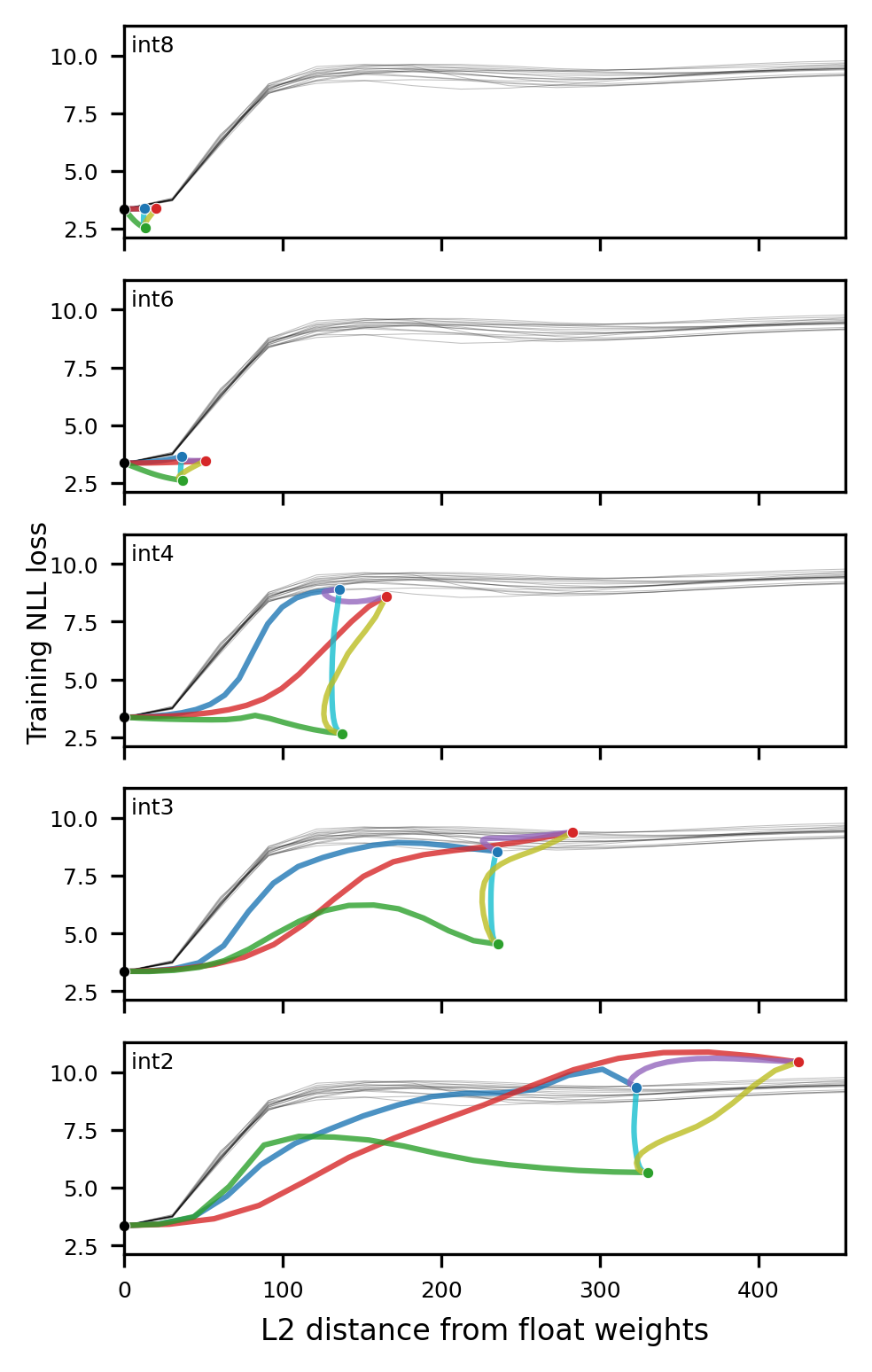}\\
\includegraphics{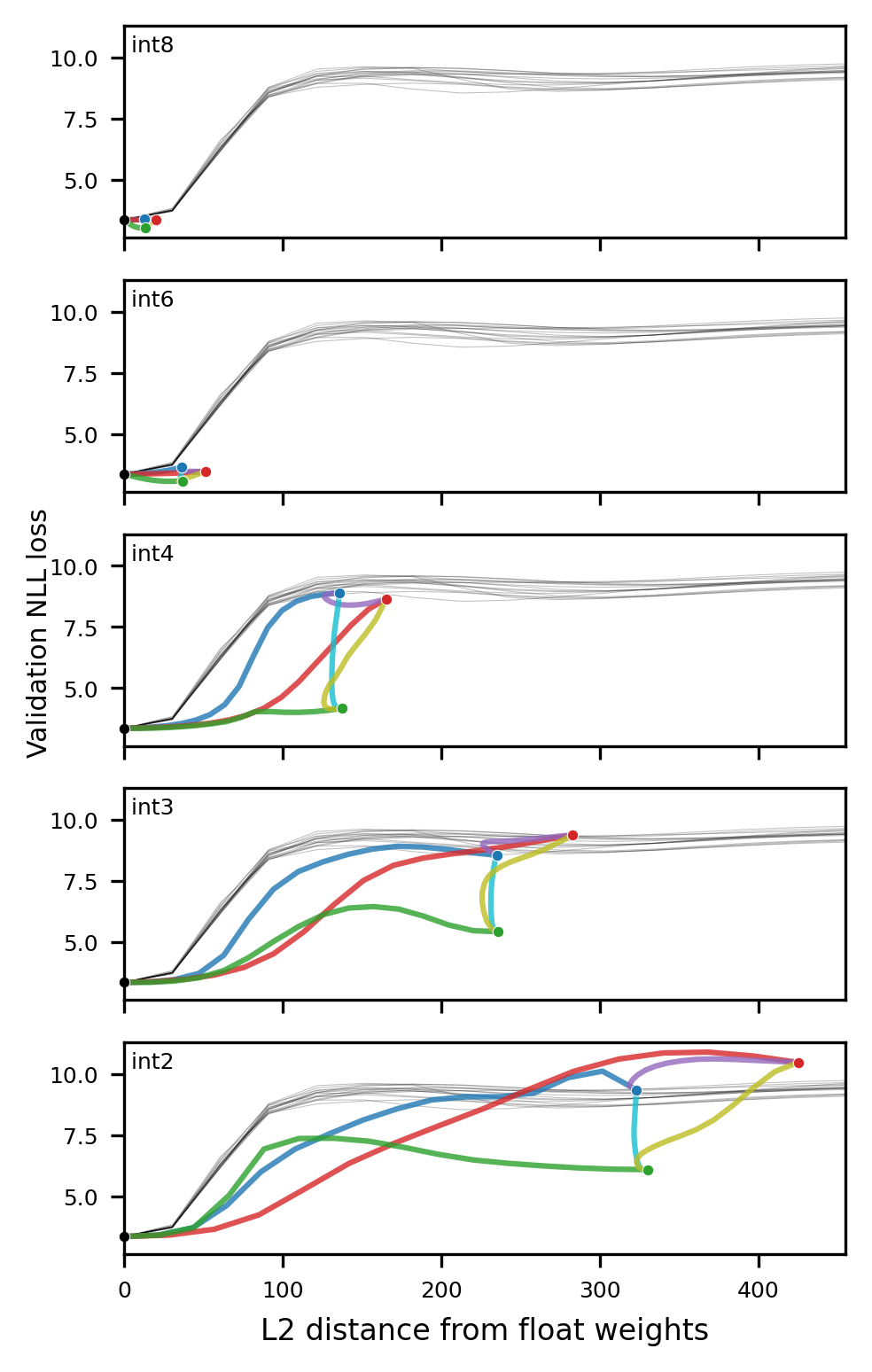}
\end{multicols}
\caption{
\textbf{Loss landscape analysis of quantized model weights.}
Data illustrated here are from \model{opt-125m}, a network small enough for numerous loss evaluations.
In the legend at the top, we illustrate the mapping strategy in a $2$-dimensional cartoon, which captures key concepts in the $D$-dimensional weight space. 
The black dot in the middle marks the pretraining convergence $\vw$. 
The continuous loss landscape is probed first by measuring loss at $\vw + \lambda \hat\ve$, \emph{i.e.} pretrained weight subject to random perturbation  $\hat\ve \sim \gS^D$ sampled uniformly from the $D$-dimensional unit sphere.
We sweep $\lambda \in \sR^+$ (thin, light gray lines emanating from the black circle) to map the radial loss landscape along a specific random direction $\hat\ve$.  
The gray grid represents the representable weight values prescribed by the weight quantizer $Q(\cdot)$, out of which we show three key quantized weights under question: $\vw_\rtn=Q(\vw)$ (blue circle), $\vw_\qaft$ (green circle), and $\vw_\gptq$ (red circle).
We measure the loss function at these key points as well as those along the linear segment resulting from a convex combination of two of these (colored lines). 
We plot the radial loss landscape ($\nll$ loss against $\ell_2$ distance from $\vw$) in the lower panels, training loss on the left and validation loss on the right.
Graphical symbols of points and segments are consistent with the legend at the top.
}
\label{fig:lls}
\vspace{0.5cm}
\end{figure*}

\subsection{GPTQ vs. QAFT under the same data constraint}

First, we investigated whether minimization of the local $\mse$ losses through $\gptq$ aligned with minimization of the global $\nll$ loss through $\qaft$.  
Both methods were applied on all experimented LLMs, as described in Section~\ref{sec:methods}, and the resulting global $\nll$ and local layer-wise $\mse$ losses were evaluated on the test dataset.

Figure~\ref{fig:gptq_vs_qaft} compares the global $\nll$ and local $\mse$ losses for \model{gpt2-xl}, \model{opt-6.7b} and \model{llama2-7b}, which were the largest models among each model family we experimented with. 
In the upper rows, global $\nll$ loss for $\qaft$ are consistently lower than those for $\gptq$, as indicated by the position of the colored dots below the diagonal identity line. 
In some cases, $\gptq$ even resulted in higher $\nll$ losses than $\rtn$. 
In contrast, the lower row shows that $\gptq$ always reduced layer-wise $\mse$ losses as designed, whereas $\qaft$ maintains or even increases the $\mse$ losses from $\rtn$. 

These results indicate a misalignment: minimizing global $\nll$ loss via $\qaft$ does not necessarily reduce local $\mse$, and minimizing local $\mse$ loss via $\gptq$ does not necessarily reduce global $\nll$. This misalignment is particularly evident in low-precision formats, where $\qaft$ significantly outperforms $\gptq$.
Since the generalization capability of a quantized model is measured by the global $\nll$ loss, minimizing layer-wise $\mse$s seems to be an ineffective surrogate, in light of the observed misalignment.  
In Appendix~\ref{app:iterations}, we showed that even a few $\qaft$ iterations can produce better generalizing quantized model than $\gptq$.

\subsection{Scaling with LLM size and numerical precision}

Next, we examined how the generalization abilities of quantized models produced by $\rtn$, $\qaft$ and $\gptq$, scaled with model size and numerical precision. 
To capture this, we plotted test $\nll$ loss against total transformer weight size (in gigabytes) for all models on a single trade-off graph (Figure~\ref{fig:tradeoff}).  

As shown in Figure~\ref{fig:tradeoff}, $\qaft$ consistently dominated the Pareto front across model families. 
For GPT-2 models, $\qaft$ quantized solutions at \INT{6} and \INT{8} occupied the Pareto front, outperforming both smaller full-precision models and larger models quantized with lower precisions. Similar trends were observed for the OPT and Llama 2 families, where $\qaft$ quantized models at \INT{4}, \INT{6}, and \INT{8} dominated.

Colored vertical strips in Figure~\ref{fig:tradeoff} highlight test $\nll$ differences between $\qaft$ and $\gptq$.
Models quantized by $\qaft$ outperforms $\gptq$ more significantly at lower precisions, such as \INT{2}, \INT{3} and \INT{4}, aligning with the earlier observation that misalignment between global $\nll$ and local 
$\mse$ losses is more pronounced at low numerical precision.


\subsection{Explanation of the misalignment from a loss landscape perspective}

\label{results:loss-landscape}

Finally, we investigated why minimization of local layer-wise $\mse$ losses did not align with minimizing the global $\nll$ loss, especially at low numerical precision.  
To address this, we analyzed the global $\nll$ loss landscape in the $D$-dimensional weight space around the pretrained weights $\vw \in \sR^D$.

We measured $\nll$ loss along a number of random directions emanating from $\vw$ (the thin, light gray lines in Figure~\ref{fig:lls}).  
The radial mapping revealed that $\vw$ sat near the bottom of an attractive basin. 
Within this near-convergence region, the loss landscape appeared quadratic, with similar profiles across various random directions.
This aligns with prior analysis of the loss Hessian spectra showing a dominant bulk subspace with a few outliers ~\citep{sagun2017eigenvalues,sagun2018empirical,ghorbani2019investigation}. Hence, a random linear combination of Hessian eigenvectors would likely stay within the bulk subspace, resulting in similar radial profiles. 
Beyond the near-convergence locality, the loss landscape deviated from a quadratic approximation and plateaued at a high level, defining the attractive basin's radius $R(\vw)$. 

Next, we charted the loss landscape of the quantized weights via $\rtn$, $\qaft$ and $\gptq$, and of linear interpolation segments between them (colored circles and line segments in Figure~\ref{fig:lls}). 
Key observations include:
\begin{itemize}
    \item $\vw_\rtn$ (blue circle): The closest quantized weight to $\vw$, It was within the attractive basin for \INT{8} and \INT{6}, near the border for \INT{4}, and outside it for \INT{3} and \INT{2}.
    \item $\vw_\gptq$ (red circle): Further from $\vw$ than $\vw_\rtn$, but along a flatter loss direction (red segment). The loss values at $\vw_\gptq$ were low for \INT{8} and \INT{6}, but high for \INT{4}, \INT{3} and \INT{2}, similar to $\vw_\rtn$ at the same precision. 
    \item $\vw_\qaft$ (green circle): Slightly further from $\vw$ than $\vw_\rtn$, yet achieving significantly lower loss level. 
    \item Connectivity: Quantized weights beyond the basin's radius were not simply connected to $\vw$. Linear interpolation between $\vw$ and these weights showed non-monotonic loss profiles with ridges in the middle.
\end{itemize}

Based on these observations, we are able to explain experimental findings from previous sections.  
Minimizing layer-wise local $\mse$ by $\gptq$ effectively identified less steeply rising directions from $\vw$, resulting in lower loss levels when $\vw_\gptq$ remained within the attractive basin (e.g. \INT{8} and \INT{6}). However, for larger quantization-induced perturbations, the shallow rise near $\vw$ is insufficient, resulting in high loss values at $\vw_\gptq$ (e.g. \INT{3} and \INT{2}).  
In contrast, $\qaft$ consistently follow less steep directions from $\vw_\rtn$, even when $\vw_\qaft$ was distance from $\vw$. This led to significantly lower loss levels, albeit in a separate attractive basin, similar to patterns observed in sparse networks~\citep{evci2020difficulty}. 

Our results suggested that the alignment between local and global loss minimization depends on the relationship between the attractive basin size $R(\vw)$ and the quantization-induced weight perturbation $\lVert \Delta\vw \rVert = \lVert \vw_\rtn - \vw \rVert$. When $\lVert \Delta\vw \rVert$ is substantially greater than $R(\vw)$, misalignment occurs.


\subsection{Generalization on different datasets}
To assess the generalizability of our findings, we repeated a subset of experiments on C4~\citep{c4} and LAMBADA~\citep{lambada} datasets. In both cases, we observed consistent misalignment between the minimization of global and local loss (Figure \ref{fig:local-global-dataset}), as well as similar trends in the loss landscape (Figure \ref{fig:lls-dataset}).

\begin{figure}[!tbh]

\centering

\includegraphics[scale=0.8]{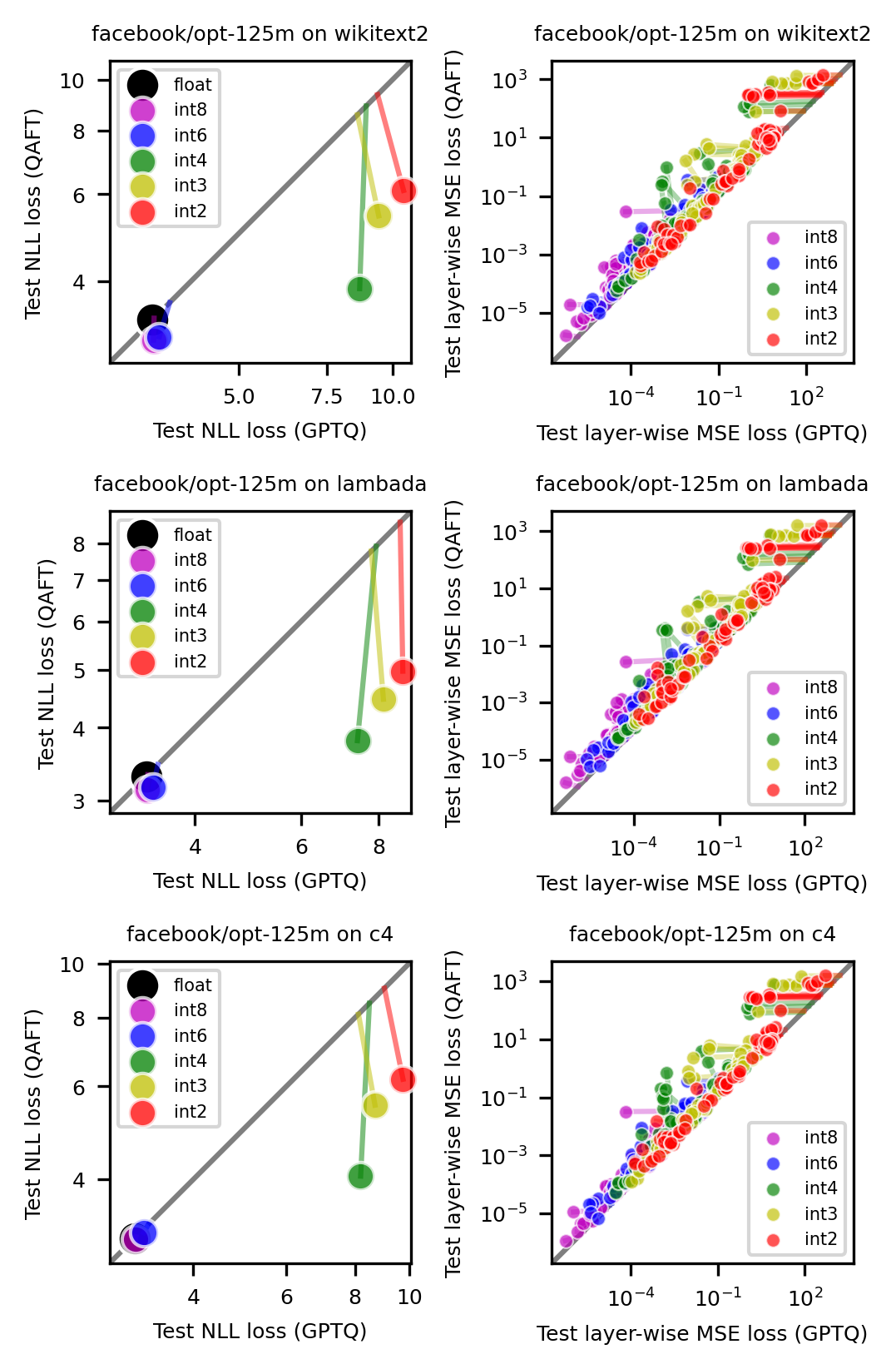} 
\caption{
\textbf{Misalignment between minimization of the global NLL loss (by QAFT) and minimization of the local layer-wise MSE losses (by GPTQ) on different datasets.} Follows the same conventions as Figure \ref{fig:gptq_vs_qaft}.
}
\label{fig:local-global-dataset}
\end{figure}

\begin{figure}[!tbh]

\centering

\includegraphics[scale=0.9]{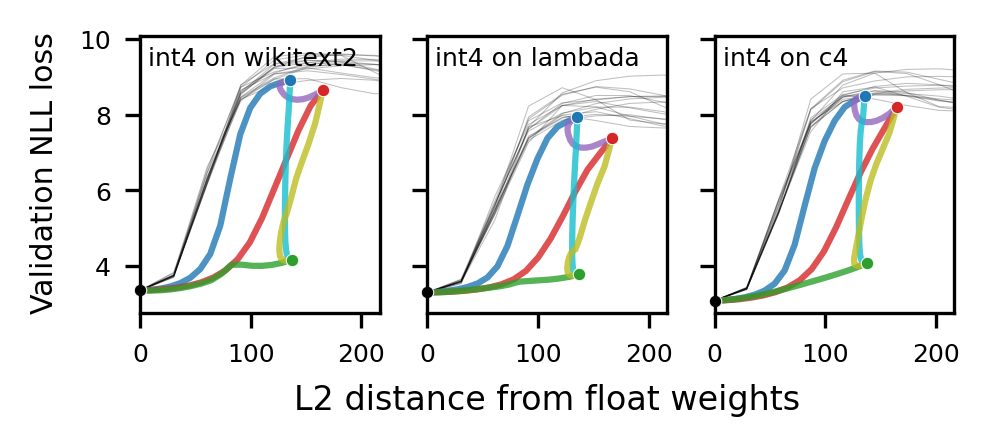} 
\caption{
\textbf{Loss landscape analysis of quantized model weights on different datasets}. Plots showing results for \model{opt-125m} quantized in \INT{4}. Follows the same conventions as Figure \ref{fig:lls}.
}
\label{fig:lls-dataset}
\end{figure}
\section{Discussion}

In this work, we systematically compared the effectiveness of post-training quantization $\gptq$ and quantization-aware fine-tuning ($\qaft$), where $\gptq$ minimized layer-wise local quantization errors and $\qaft$ minimized the global training loss. 
Under the same low training data constraint ($128$ training examples), $\qaft$ consistently outperformed $\gptq$.

Surprisingly, we observed that $\gptq$'s local $\mse$ minimization and $\qaft$'s global $\nll$ minimization often misalign, especially pronounced at very low precision quantization.  
Through loss landscape analysis, we elucidated that such misalignment was prominent when the weight perturbation due to quantization was significantly larger than the radius of the attractive basin at the pretraining convergence.   

Our findings reveal a lack of correlation between a quantized network's generalization ability and its local quantization errors, challenging the common reliance on local error minimization metrics for evaluating quantization schemes. We urge caution in generalizing the utility of local-error-based post-training quantization and provided a new perspective for understanding the difficulty and identification of conditions where these methods are effective.

\bibliography{references}

\appendix

\subsection{Optimal learning rate for $\qaft$}

While performing hyperparameter grid search on initial learning rate over the set $\{10^{-6},10^{-5},10^{-4},10^{-3}\}$, we discovered that the optimal choices strongly correlated with the model size and the quantization precision, which is supported by Fig \ref{fig:lr}.
For a particular model, we observe that $\qaft$ at higher precision such as \INT{8}, \INT{6} and \INT{4} favored low learning rates, whereas $\qaft$ at lower precision such as \INT{3} and \INT{2} preferred high learning rates. 
For a specific format, larger models in the same model family preferred lower learning rates while smaller models in the model family prefers higher learning rates. 
\label{app:lr}
\begin{figure}[!tbh]
\centering
\includegraphics[scale=0.4]{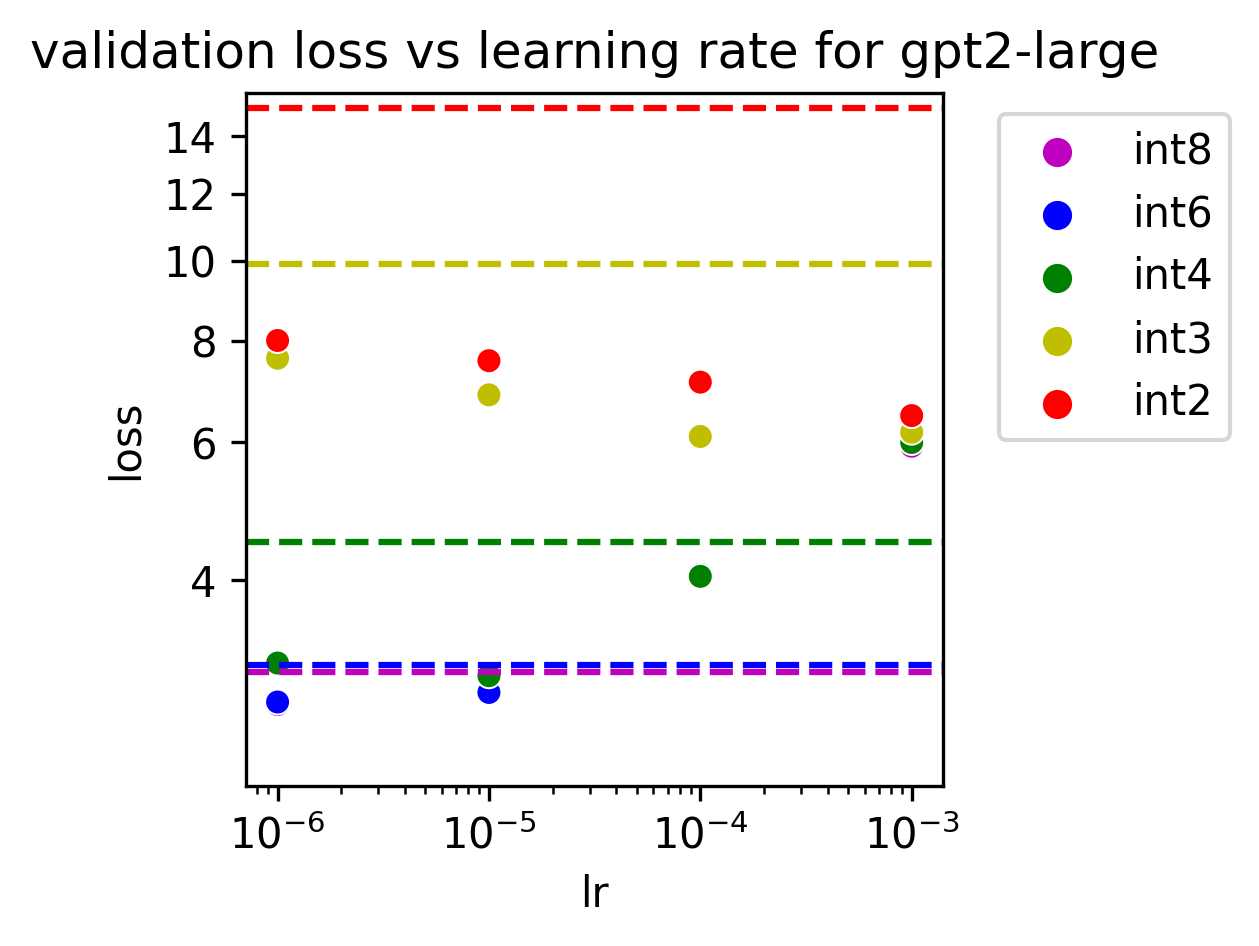}
\includegraphics[scale=0.3]{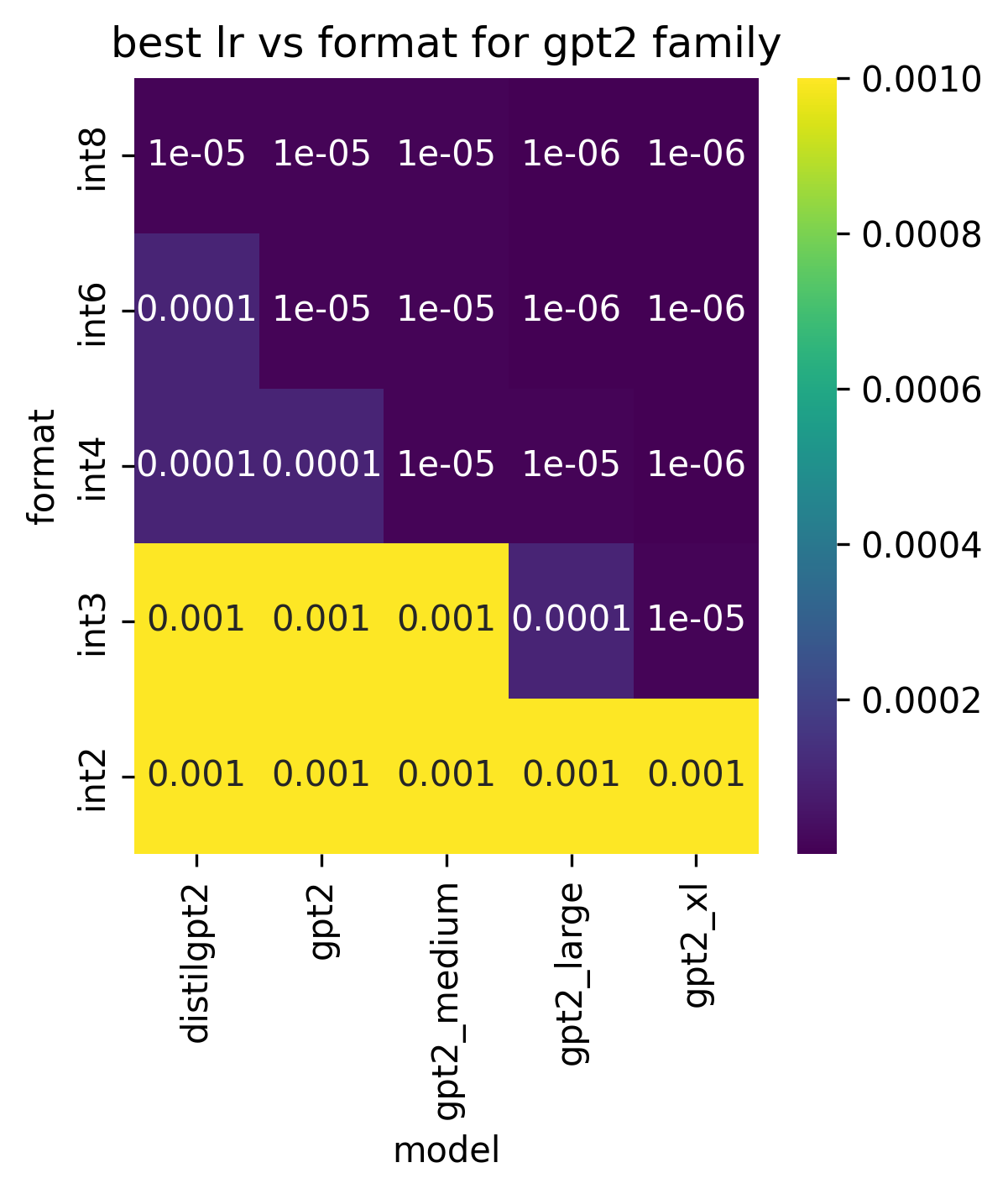}
\caption{Left: validation $\nll$ loss of \model{gpt2-large} after fine-tuning.  Dashed lines represent the loss after $\rtn$.  Right: best learning rates for GPT-2 family.}
\label{fig:lr}
\end{figure}
\vspace*{10mm}

\subsection{Effect of number of $\qaft$ iterations}
\label{app:iterations}
Although we used 8 epochs ($1024$ iterations) to achieve the most optimal performance, we discovered that $\qaft$ for $1$ epoch is sufficient to outperform $\gptq$, as shown in Figure~\ref{fig:num_epoch}.
\begin{figure} [!tbh]
\centering
\includegraphics[scale=0.35]{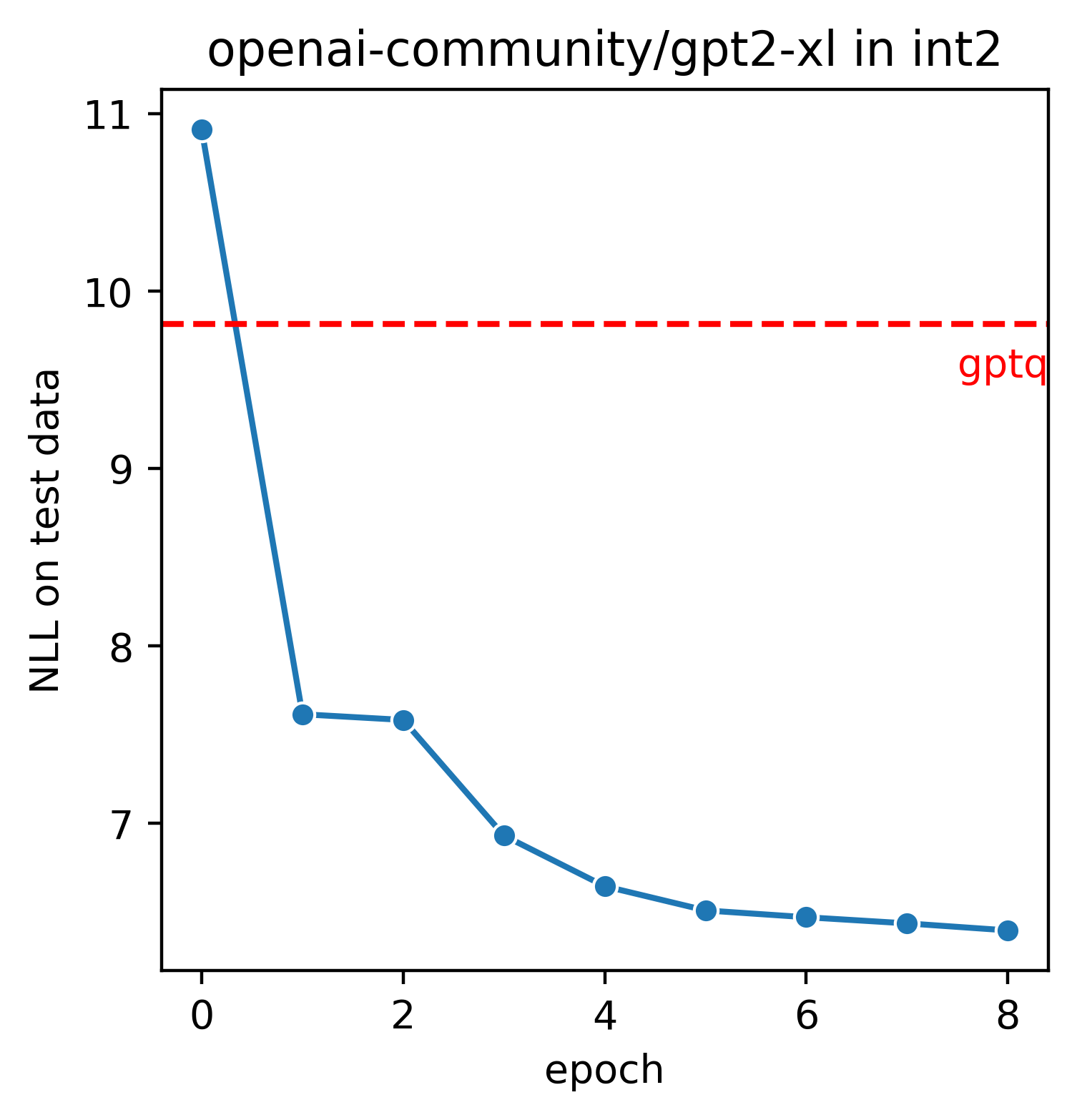}
\includegraphics[scale=0.35]{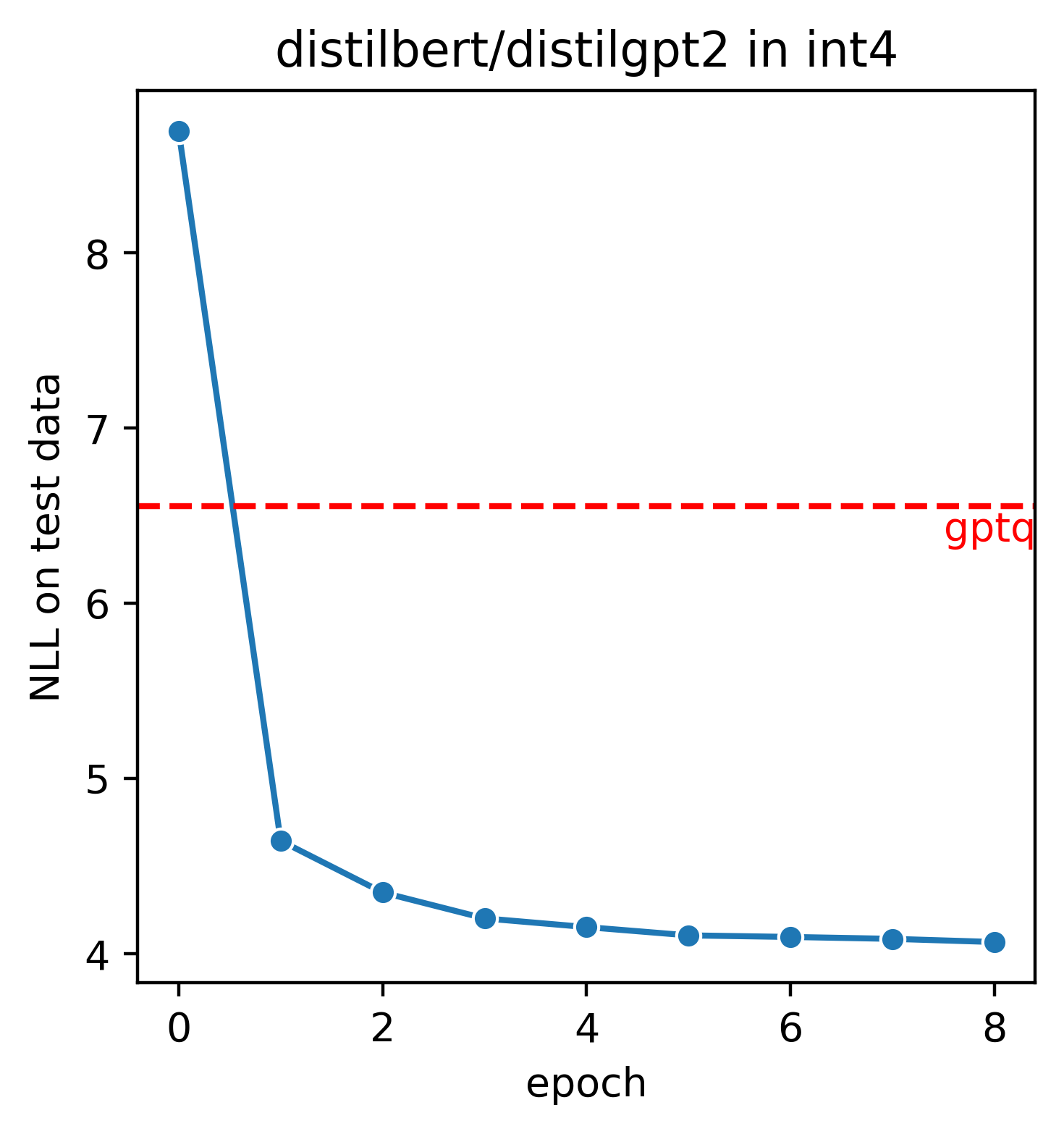}
\caption{$\nll$ on test data after each epoch of fine-tuning.  Epoch $0$ represents the $\nll$ of $\rtn$, which is the starting point for fine-tuning.  The horizontal red line marks the $\nll$ after $\gptq$.}
\label{fig:num_epoch}
\end{figure}

\subsection{Optimal $\gptq$ dampening factor}
\label{app:percdamp}
In $\gptq$, dampening factor is multiplied with the average diagonal value in Hessian matrix $H$ and added to the diagonal entries of $H$ to achieve better numerical stability \cite{frantar2022gptq}. In our experiments, we did a grid search on dampening factor for each layer with the objective of minimizing layer-wise $\mse$. We visualized the best dampening factor for each quantized layer but failed to find any meaningful patterns.

\end{document}